\documentclass[10pt,twocolumn,letterpaper]{article}

\usepackage{iccv}
\usepackage{times}
\usepackage{epsfig}
\usepackage{graphicx}
\usepackage{amsmath}
\usepackage{amssymb}

\usepackage{algorithm}
\usepackage[noend]{algpseudocode}
\usepackage{stfloats}
\usepackage{caption}
\usepackage{setspace}
\usepackage{lipsum}
\usepackage{multirow}
\usepackage{diagbox}
\usepackage{bm}
\usepackage{mathrsfs}

\usepackage{subfigure}
\makeatletter 
\renewcommand{\@thesubfigure}{\hskip\subfiglabelskip}
\makeatother

\usepackage[breaklinks=true,bookmarks=false]{hyperref}
\iccvfinalcopy

\def\iccvPaperID{1075} 

\ificcvfinal\fi

\begin{document}

\title{StarEnhancer: Learning Real-Time and Style-Aware Image Enhancement}

\renewcommand{\thefootnote}{\fnsymbol{footnote}}
\author {
	Yuda Song \textsuperscript{\rm 1}
	\quad Hui Qian \textsuperscript{\rm 1}
	\quad Xin Du \textsuperscript{\rm 2}\footnotemark[1]\\
	\textsuperscript{\rm 1}College of Computer Science and Technology, Zhejiang University, Hangzhou, China\\
	\textsuperscript{\rm 2}College of Information Science and Electronic Engineering, Zhejiang University, Hangzhou, China \\
	{\tt\small \{syd,qianhui,duxin\}@zju.edu.cn}
}

\maketitle

\footnotetext[1]{Corresponding author.}
\renewcommand{\thefootnote}{\arabic{footnote}}

\ificcvfinal\thispagestyle{empty}\fi

\begin{abstract}
    Image enhancement is a subjective process whose targets vary with user preferences.
    In this paper, we propose a deep learning-based image enhancement method covering multiple tonal styles using only a single model dubbed StarEnhancer.
    It can transform an image from one tonal style to another, even if that style is unseen.
    With a simple one-time setting, users can customize the model to make the enhanced images more in line with their aesthetics.
    To make the method more practical, we propose a well-designed enhancer that can process a 4K-resolution image over 200 FPS but surpasses the contemporaneous single style image enhancement methods in terms of PSNR, SSIM, and LPIPS.
    Finally, our proposed enhancement method has good interactability, which allows the user to fine-tune the enhanced image using intuitive options.
\end{abstract}

\section{Introduction}

The development of smartphone cameras has dramatically lowered the barriers to take photos, but amateurs still lack the skills to get high-quality photos.
To this end, a variety of image post-processing techniques have been proposed to bridge the gap.
Generally, these techniques tend to improve the quality of image detail, which is broadly objective.
However, the quality of a photo depends not only on the image detail but also on whether the photo meets people's aesthetics, which is entirely subjective.
Therefore, various deep learning-based approaches~\cite{chen2018deep,guo2020zero,jiang2021enlightengan} are proposed to retouch photos to make these photos more aesthetically pleasing.
But only a few works~\cite{gharbi2017deep,kim2020pienet} have realized the difference between image enhancement and other low-level vision tasks.
In our opinion, a practical image enhancement method should not employ a generic image restoration network but aim to be real-time and style-aware.

\begin{figure}[t]
    \centering
    \includegraphics[width=1.0\columnwidth]{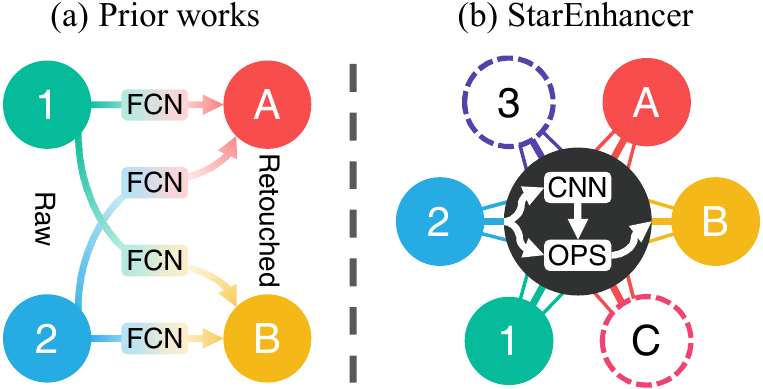}
    \caption{
        Comparison between prior works and our proposed method.
        (a) Prior works need to train multiple models to learn the mappings between source styles and target styles.
        (b) StarEnhancer offers the capability to transform images from one style to another using a single model.
    }
    \label{fig:motivation}
\end{figure}

It is straightforward that different users have different aesthetic preferences, so the target image's tonal style is not constant~\cite{hu2018exposure,kim2020pienet,kneubuehler2020flexible}.
And different cameras have different camera response functions (CRFs) and image signal processing (ISP) pipeline~\cite{afifi2020deep,grossberg2003determining}, which means that the tonal style of the input image is also not constant.
Meanwhile, users may want to transform the retouched images into their preferred style, whereas the unretouched images are not available.
Therefore, we consider that a practical image enhancement method requires the capacity to transform images between multiple tonal styles.
Finally, because user preferences are challenging to quantify precisely, the intuitive manual adjustment options are appealing~\cite{he2020conditional,li2020flexible}.

The current common image restoration strategy is to train a fully convolutional network (FCN)~\cite{long2015fully} to reconstruct the images fed into the network.
However, the computational complexity of the FCN grows quadratically with the spatial dimensions of the input images~\cite{howard2017mobilenets}.
And the FCN-based network is more difficult to train and may introduce artifacts~\cite{gharbi2017deep}, especially when employing generative adversarial networks (GANs)~\cite{bickel2020multiple,chen2018deep,goodfellow2014generative}.
Besides, the FCN's receptive field does not change with the input image's size, which may lead to a visible dissimilarity between the enhanced images of the same image in different resolutions~\cite{gharbi2017deep}.
In contrast, color transform-based image enhancement methods~\cite{gharbi2017deep,zeng2020learning} only use convolutional neural networks (CNNs) to encode the color transformation's parameters from a fixed-size, low-resolution version of the image.
And the learned color transformation functions can be applied to the full-resolution images, whose computational complexity is extremely low.
Therefore, we consider the color transform-based image enhancement methods may be plausible solutions.

As illustrated in Figure~\ref{fig:motivation} (a), most existing image enhancement methods need to train an individual FCN for each transformation.
More critically, a new dataset needs to be collected for each new style transformation, yet collecting a dataset for image enhancement is challenging since it relies on expert knowledge~\cite{bychkovsky2011learning,wang2019underexposed}.
Also, considering the high computational complexity of FCN when processing high-resolution images, such methods can only provide limited capabilities to users.

To this end, we propose a more practical image enhancement method that is real-time and style-aware to bridge these gaps.
We named our method StarEnhancer to admire StarGAN~\cite{choi2018stargan,choi2020stargan}, albeit our approach is vastly different from StarGAN.
As demonstrated in Figure~\ref{fig:motivation} (b), StarEnhancer utilizes multiple tonal styles' training data and learns the mapping between multiple tonal styles using a single model.
Specifically, we first train a style classifier to classify images and take the output embedding vector of the classifier's penultimate layer as the latent codes.
The mapping network then encodes these latent codes as a set of style codes, which customizes the curve encoder using Dual AdaIN modified from adaptive instance normalization (AdaIN)~\cite{huang2017arbitrary}.
The curve encoder predicts the curves' parameters from the low-resolution version of the image, and the enhancer applies these curves to transform the full-resolution image.
Unlike the existing color transform-based image enhancement methods~\cite{chai2020supervised,gharbi2017deep,kim2020global,kneubuehler2020flexible,liu2020color,moran2021curl,wang2019underexposed,zeng2020learning}, StarEnhancer considers the correlation between color channels and the pixel's coordinates.

\vspace{0.25 cm}
Overall, our contributions are as follows:

\begin{itemize}
    \item We propose a highly efficient curve-based enhancer that can enhance a 4K-resolution image over 200 FPS on a single GPU.
    And it is scale-invariant and artifact-free, which is critical for high-resolution images.
    \item We propose a flexible approach named StarEnhancer for image enhancement between multiple styles.
    It can be customized to meet different camera characteristics and user preferences via a simple one-time setup.
    \item StarEnhancer provides intuitive options to allow users to fine-tune the results for each image manually.
    \item StarEnhancer achieves state-of-the-art performance in terms of efficiency and effectiveness on the MIT-Adobe-5K dataset~\cite{bychkovsky2011learning}.
\end{itemize}

\section{Basic enhancer}

The effectiveness and efficiency of the image enhancer can substantially affect the approach's practicality.
Therefore, we first discuss how to design an expressive and fast image enhancer.
Finally, we propose the basic enhancer for the single style transformation.

\subsection{Problem formulation}

Unlike the real-time image classification methods that emphasize network architecture design~\cite{howard2017mobilenets,tan2019efficientnet,zhang2018shufflenet}, the real-time image restoration methods also involve the impact of the input image's size on the processing efficiency.
Deep learning-based image restoration methods are generally based on the FCNs, making their computational complexity quadratic to the input image's spatial dimensions.
And most image enhancement methods also employ FCN-like network architectures~\cite{afifi2021learning,chen2017fast,chen2018deep,he2020conditional,kinoshita2019convolutional,zamir2020learning}.
If using the network $\mathbf{G}$ with parameters $\theta$ to enhance the input image $I$ directly, the output image $O$ can be formulated as follows:
\begin{equation}
    O = {\mathbf{G}}(I;\theta).
\end{equation}

Fortunately, the standard image enhancement task is more aware of the global information, making it possible to utilize the down-sampled image to obtain informative features, just like the strategy applied in the high-level vision task.
As a trade-off, it isn't easy to apply this kind of method to the detail-concerned image restoration tasks, including even the low-light image enhancement task~\cite{liang2020deep,lore2017llnet,tao2017llcnn,wei2018deep,zhang2019kindling}.
Specifically, the network $\mathbf{G}$ with parameters $\theta$ extracts the features from the down-sampled input image $I{\downarrow}$, and these features are used to formulate the transformation function $\mathbf{F}$ applied to the input image $I$ as follows:
\begin{equation}
    O = \mathbf{F}(I;{\mathbf{G}}({I{\downarrow}};\theta)).
\end{equation}

In this way, the backbone network's computational complexity for extracting features hardly varies with the input image size.
And the key to designing a powerful image enhancer is to develop an efficient and expressive transformation function.

\subsection{Prior art}

There are roughly three categories of functions to follow: color transformation matrix~\cite{chai2020supervised,gharbi2017deep,liu2020color}, curve-based color transformation function~\cite{guo2020zero,kim2020global,li2020flexible,moran2021curl}, and 3D lookup table (LUT)~\cite{zeng2020learning}.

The color transformation matrix is a $3 \times 4$ affine transformation matrix that maps the pixel's input color to the output color.
As an example, HDRNet~\cite{gharbi2017deep} predicts low-resolution affine color transformation coefficient matrices in the bilateral grid.
Guided by the full-resolution single-channel guide map, these matrices are sliced into full-resolution coefficient matrices that are then applied to the original image.
However, the color transformation matrix's capability is not sufficient.
And the guide map is generated by an FCN built with several point-wise convolution layers, which is still expensive for high-resolution images.

Curve-based color transformation functions mimic the color adjustment curve tool in retouching software (\emph{e.g.} Lightroom and Photoshop) and are more in line with human retouching.
In order to quantify a curve using a limited number of parameters, the backbone networks regress the knot points of the curve~\cite{kim2020global,li2020flexible,moran2021curl} or the coefficients of a pre-defined function (\emph{e.g.} gamma function, polynomial function)~\cite{guo2020zero,hu2018exposure,park2018distort}.
However, existing curve-based color transformation functions mostly neglect the relationship between color channels, which is in line with the color adjustment curve tool's characteristics but lacks the capability to approximate complex transformations.

3D LUTs are expressive operators that have been used in ISP~\cite{karaimer2016software}.
Generally, 3D LUTs are obtained by expert adjustment and are fixed after the adjustment.
To this end, Adaptive 3DLUT~\cite{zeng2020learning} learns a set of basis 3D LUTs from the training dataset and uses a CNN to predict content-dependent weights from the down-sampled image.
These weights are used to fuse multiple basis 3D LUTs into a single 3D LUT that is then applied to the full-resolution image's transformation.
Sincerely, 3D LUT is an attractive transformation function.
However, Adaptive 3DLUT is a trade-off approach because only the weights for fusing the basis 3D LUTs are adaptive to the input image, while the basis 3D LUTs are still fixed after training.
We believe that it is not flexible enough to be used as the basic image enhancer for multi-style image enhancement.

\subsection{Multi-curve enhancer}

\begin{figure}[t]
    \centering
    \includegraphics[width=1.0\columnwidth]{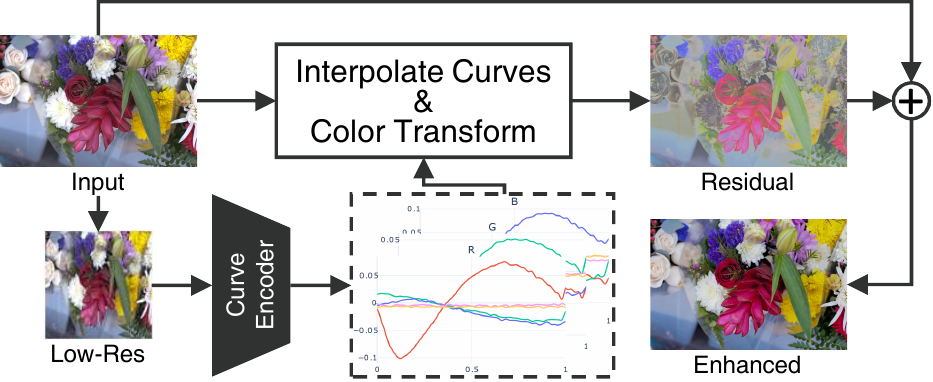}
    \caption{
        Framework of the proposed basic image enhancer.
    }
    \label{fig:basic}
\end{figure}

We consider building our enhancer based on a curve-based color transformation function.
For the prior curve-based image enhancement methods, the color transformation for output channel ${\rm{j}} \in \{ {\rm{r,g,b}}\}$ can be formulated as:
\begin{equation}
    {O_{\rm{j}}} = {\bf{F}}(I_{{\rm{j}}};{\bf{G}}{({I{\downarrow}};\theta )_{\rm{j}}}).
\end{equation}

Combined with the ideas of color transformation matrix and 3D LUT, we propose a revised curve-based transformation function to build our basic enhancer.
We note that both the color transformation matrix and the 3D LUT take into account the correlation between the color channels. 
In other words, each channel of the output image is correlated with each channel of the input image.
Besides, we believe that introducing the coordinate maps $\{ \rm{x,y} \}$ can make the transformation function more expressive.
Thus the revised transformation for input pixel $I(x,y)$ ($I,x,y \in [0,1]$) can be formulated as:
\begin{equation}
    {O_{\rm{j}}}(x,y) = {\bf{F}}(I(x,y),x,y;{\bf{G}}{({I{\downarrow}};\theta )_{\rm{j}}}).
\end{equation}

Figure~\ref{fig:basic} illustrates how our proposed image enhancer processes a high-resolution image with $H \times W$ resolution.
Firstly, the CNN-based curve encoder predicts a parameter vector $\bf{u}$ for all curves' knot points from the down-sampled input image with $K \times K$ resolution.
Then we split $\bf{u}={\mathbf{G}}({I{\downarrow}};\theta)$ into $15$ subvectors, in which ${\bf{u}}_{\rm{i,j}}$ corresponds to the curve that maps input channel ${\rm{i}} \in \{ {\rm{r,g,b,x,y}}\}$ to output channel ${\rm{j}} \in \{ {\rm{r,g,b}}\}$.
We propose an extremely fast curve-based transformation using piecewise cubic interpolation~\cite{fritsch1980monotone} and indexing to utilize the knot points' parameter vectors.
Using the piecewise cubic interpolation function $\mathcal{S}^{M,N}$, we interpolate an $M$-dimensional vector ${{\bf{u}}_{\rm{i,j}}} = [{u_{{\rm{i,j}},0}},...,{u_{{\rm{i,j}},M-1}}]^T$ to an $N$-dimensional vector ${{\bf{v}}_{\rm{i,j}}} = [{v_{{\rm{i,j}},0}},...,{v_{{\rm{i,j}},N-1}}]^T$ as follows:
\begin{equation}
    {\bf{v}}_{\rm{i,j}} = \mathcal{S}^{M,N}({\bf{u}}_{\rm{i,j}}).
\end{equation}
Let ${\bf{v}}_{\rm{i,j}}(k)$ be the $k$-th element ${v}_{{\rm{i,j}},k}$ of ${\bf{v}}_{\rm{i,j}}$, we apply following transformation to obtain the residual image $R$:
\begin{equation}
    \begin{split}
        {R_{\rm{j}}}(x,y)   & = \mathcal{S}^{{M_{\rm{r,j}}},2^D}({\bf{u}}_{\rm{r,j}}) (\left\lfloor {{I_{\rm{r}}}(x,y) \cdot (2^D - 1)} \right\rfloor ) \\
                            & + \mathcal{S}^{{M_{\rm{g,j}}},2^D}({\bf{u}}_{\rm{g,j}}) (\left\lfloor {{I_{\rm{g}}}(x,y) \cdot (2^D - 1)} \right\rfloor ) \\
                            & + \mathcal{S}^{{M_{\rm{b,j}}},2^D}({\bf{u}}_{\rm{b,j}}) (\left\lfloor {{I_{\rm{b}}}(x,y) \cdot (2^D - 1)} \right\rfloor ) \\
                            & + \mathcal{S}^{{M_{\rm{y,j}}},H}({\bf{u}}_{\rm{y,j}}) (\left\lfloor {y \cdot (H - 1)} \right\rfloor ) \\
                            & + \mathcal{S}^{{M_{\rm{x,j}}},W}({\bf{u}}_{\rm{x,j}}) (\left\lfloor {x \cdot (W - 1)} \right\rfloor ),
    \end{split}
\end{equation}
where $\left\lfloor \cdot \right\rfloor$ is the floor function and $D$ denotes the color depth of each channel.
In practice, we only require the interpolated vector $\{ {\bf{v}}_{\rm{i,j}} \} _{{\rm{i}} \in \{ \rm{x,y} \}}$  to be expanded to a map with the same resolution as the input image since the coordinates of the pixels are monotonic.
Besides, it is feasible to apply a low color depth transformation to a high color depth image to reduce the cost of indexing (\emph{e.g.} $D=8$ for 48-bit color image).
In this way, we need to render the residual image instead of rendering the enhanced image for preserving the information of high color depth images.
Finally, the enhanced image can be obtained by $O = R + I$.

Given the image pair $\{ I_a,I_b \}$, where $I_a$ is the input image and $I_b$ is the reference image, we compute the $L_1$ loss in CIELab color space to train the enhancer:
\begin{equation}
    {\mathcal{L}_E} = {\left\| {Lab(I_b) - Lab({\mathbf{F}(I_a;{\mathbf{G}}({I_a {\downarrow}};\theta ))})} \right\|_1}.
\end{equation}

\section{StarEnhancer}

In this section, we illustrate how to guide the enhancer to perform adaptive multi-style color transformations.
There are two critical issues to be addressed: how to make the method adaptive to unseen styles and how to feed style information into the network.

\subsection{Style encoder}

\begin{figure}[t]
    \centering
    \includegraphics[width=1.0\columnwidth]{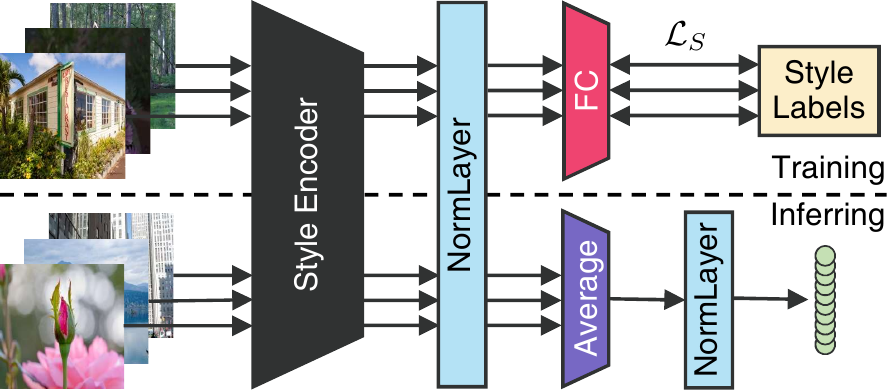}
    \caption{
        Framework of our proposed style encoder.
        When inferring, the input images are all from the same style.
    }
    \label{fig:stylish}
\end{figure}

We first discuss how to design the style code to encode unseen styles, thereby making the enhancer adaptive to new cameras and users.
Apparently, a fixed label such as the one-hot vector used in StarGAN~\cite{choi2018stargan} is not a good choice.
In contrast, the latent code used in StarGAN v2~\cite{choi2020stargan} is a better choice, but it is obtained by randomly sampling from a known distribution, which only ensures diversity but does not establish a clear link to the style.
Inspired by face recognition works~\cite{deng2019arcface,liu2017sphereface,wang2017normface,wang2018cosface}, we strive to train a style encoder that can learn image embeddings to establish the link between the specific style and the latent code.

Figure~\ref{fig:stylish} illustrates an overview of our proposed approach.
Specifically, we first train an image classifier on a dataset containing images of multiple tonal styles.
Given the embedding $\bf{f}$ of the downsampled input image after the final global pooling layer and the corresponding style class label $p$, the loss can be formulated as follows:
\begin{equation}
    {\mathcal{L}_S} = -{{\rm{log}}\left(
         {\frac
         {{{\rm{exp}}\left( \frac{{{{\bf{f}}^T}{{\bf{w}}_p}}}{{{{\left\| {\bf{f}} \right\|}_2}{{\left\| {{{\bf{w}}_p}} \right\|}_2}}} \cdot s \right)}}
         {\sum\nolimits_{q \in Q} {{\rm{exp}}\left( {\frac{{{{\bf{f}}^T}{{\bf{w}}_q}}}{{{{\left\| {\bf{f}} \right\|}_2}{{\left\| {{{\bf{w}}_q}} \right\|}_2}}} \cdot s} \right)} }
         } 
    \right)},
\end{equation}
where $s$ is a scaling term, $Q$ denotes the style classes set, $\bf{w}$ is the weight of the last fully connected layer without bias term, and ${{\left\| \cdot \right\|}_2}$ is the $L^2$-norm.

In the inference phase, we feed $n$ images of the specific style into the style encoder and obtain the embeddings $\{ {{\bf{f}}_{i}} \} _{i=1,...,n}$ after the global pooling layer.
We approximate the embedding of the specific style by calculating the average of the $L^2$-normalized embeddings:
\begin{equation}
    {{\bf{f}}_{{\rm{avg}}}} = \frac{1}{n}\sum\limits_{i = 1}^n {\frac{{{{\bf{f}}_i}}}{{{{\left\| {{{\bf{f}}_i}} \right\|}_2}}}}.
    \label{eq:avg}
\end{equation}
Since the average embedding does not always fall on the unit sphere as the single $L^2$-normalized embedding, we also apply $L^2$-normalization to it as follows:
\begin{equation}
    {{\bf{\tilde f}}} = \frac{{{{\bf{f}}_{{\rm{avg}}}}}}{{{{\left\| {{{\bf{f}}_{{\rm{avg}}}}} \right\|}_2}}}.
    \label{eq:renorm}
\end{equation}
We treat ${{\bf{\tilde f}}}$ as the center embedding of the specific style, as well as the latent code of the style.

\subsection{Multi-style enhancer}

\begin{figure}[t]
    \centering
    \includegraphics[width=1.0\columnwidth]{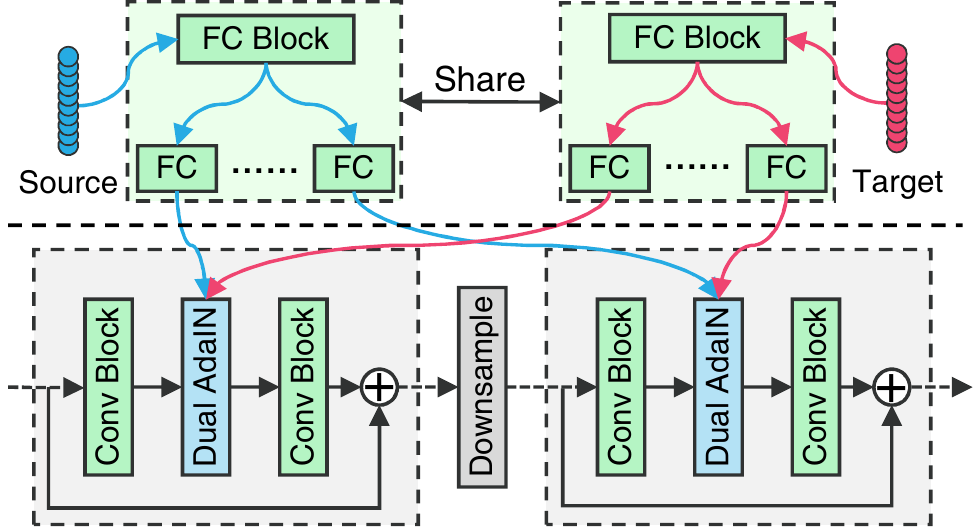}
    \caption{
        Framework of feeding the latent codes into the curve encoder.
        The top is the mapping network, while the bottom is the curve encoder with Dual AdaINs.
    }
    \label{fig:feed}
\end{figure}

Since the style-specific latent code has been obtained, now we need to feed the latent code into the curve encoder.
We suppose that image enhancement can be viewed as a special type of style transfer so that AdaIN~\cite{huang2017arbitrary} may be a choice worth employing.
However, we find that the normalization layer in the convolution block always leads to poor performance.
To this end, we propose Dual AdaIN, which does not calculate the mean and variance of the feature maps for each input sample, but the mean and variance are obtained by mapping the latent codes to the style codes.

Figure~\ref{fig:feed} illustrates how to extend the basic enhancer to StarEnhancer.
Firstly, we inserted the Dual AdaIN into the curve encoder of the basic enhancer in Figure~\ref{fig:basic}.
Then we fetch the latent codes $\{ {{\bf{\tilde f}}_d} \} _{d \in \{ {a,b} \}}$ of the source style class $a$ and the target style class $b$ using the style encoder.
Given the latent codes $\{ {{\bf{\tilde f}}_d} \} _{d \in \{ {a,b} \}}$, the mapping network maps them to $L$ sets of style codes $\{{ {\mu}_{d,1},{\sigma}_{d,1},\ldots,{{\mu}_{d,L},{\sigma}_{d,L}} }\} _{d \in \{ {a,b} \}}$, which are then fed into the curve encoder via Dual AdaIN:
\begin{equation}
    {{\mathcal{F}}_{j}'} = {\sigma _{b,j}}\left( {\frac{{{{\mathcal{F}}_{j}} - {\mu _{a,j}}}}{{{\sigma _{a,j}}}}} \right) + {\mu _{b,j}},
\end{equation}
where $j \in \{ {1,\ldots,L} \} $, ${\mathcal{F}}$ is the input feature map, and ${\mathcal{F}}_{j}'$ is the transformed feature map.

We also compute the $\mathcal{L}_E$ loss to train the multi-style enhancer, except that the training pair $\{ I_a,I_b,{\bf{\tilde f}}_a,{\bf{\tilde f}}_b \}$ is randomly sampled from all possible styles ($a,b \in Q$).

\subsection{User awareness}

If the curve encoder and mapping network are trained using only the center embeddings of specific styles in the train set, they may tend to overfit these embeddings.
To this end, we use subsets of the train set to generate more style embeddings, \emph{i.e.}, feed fewer images of the specific style to generate additional style embeddings via Eq.(\ref{eq:avg}) and Eq.(\ref{eq:renorm}).

New users can select their preferred images in the shared gallery or use their collection to generate new target latent codes.
And the latent code of the source style can be pre-generated by the camera manufacturer or obtained using several unretouched images.
Note that paired images are not necessary for this procedure.

We further provide manual fine-tuning options, which is very useful when the results do not meet user preference.
For experts, all knot points of the predicted curves can be adjusted manually, just like the curve tool in Lightroom.
But such a curve tool is still too difficult for non-experts, so we further propose a slider-based manual fine-tuning tool.
Specifically, the user can adjust the sliders that correspond to $\{ {\beta_{\rm{i,j}}} \}$ for tuning the contribution of each curve:
\begin{equation}
    {{\bf{u'}}_{\rm{i,j}}} = {\beta_{\rm{i,j}}}  \cdot {{\bf{u}}_{\rm{i,j}}}.
\end{equation}
We use $\{ {\bf{u'}}_{\rm{i,j}} \}$ to generate new curves and apply them to transform the image.
Because our proposed curve-based enhancer is highly efficient and the manual fine-tuning procedure does not perform CNN inference, users can get feedback in real-time and further adjust the sliders.

\subsection{Implementation detail}

We build our StarEnhancer using PyTorch~\cite{paszke2019pytorch}, and all operations used in the enhancer have been efficiently and differentiably implemented.
Both the style encoder and the curve encoder are built on shallow ResNet~\cite{he2016deep}, but with all batch normalization layers~\cite{ioffe2015batch} removed from the convolution blocks.
For a stable training after removing the batch normalization layers, we apply the Fixup initialization~\cite{zhang2018fixup} as well as a few architecture modifications to the network.

We first train the style encoder using $\mathcal{L}_S$ loss and obtain the latent code ${{\bf{\tilde f}}_{q}}$ for each style class $q$, and then train the mapping network and curve encoder using ${\mathcal{L}_E}$ loss.
When training the style encoder, we set the scaling term $s$ in $\mathcal{L}_S$ loss to a large constant and assign a much larger learning rate to the last fully connected layer.
All models are trained using the Adam optimizer~\cite{kingma2015adam} with the cosine annealing strategy~\cite{loshchilov2016sgdr} but not warm restarts.

When inferring, the style encoder and mapping network are only performed only once initially, then the fetched style codes are stored for future use.
Further, the users can upload the preferred images to the server, which returns the corresponding style codes, so that the user devices only need to keep the model weights for the curve encoder.

\section{Experiments}

\subsection{Experimental setup}

We train and evaluate our method on the MIT-Adobe-5K dataset~\cite{bychkovsky2011learning}, which is the only dataset that consists of images in multiple expert retouching styles.
MIT-Adobe-5K contains 5000 images captured by DSLR, each corresponding to a total of 12 styles, including 5 expert retouching styles (Artist A/B/C/D/E), 4 camera input styles, and 3 auto-retouching styles.
StarEnhancer is the first method that exploits all the data in the MIT-Adobe-5K to the best of our knowledge.

\textbf{Single style enhancement:}
we follow the experimental setup of the MIT-Adobe-5K-UPE benchmark~\cite{wang2019underexposed} to evaluate our method's performance.
Specifically, we use the images of the default input style as input, the images retouched by Artist C as the ground truth, and split the dataset into 4500 training image pairs and 500 test image pairs.
All images in the test set retain their original resolution, varying from $2160 \times 1440$ to $6048 \times 4032$.
We evaluate our method quantitatively using PSNR, SSIM, and LPIPS~\cite{zhang2018unreasonable} to compare with contemporaneous methods.

\textbf{Multi-style enhancement:}
we extend the experimental setup of the MIT-Adobe-5K-UPE benchmark to 10 styles, including 5 expert retouching styles (A/B/C/D/E), 3 camera input styles (O/P/Q), and 2 auto-retouching styles (X/Y).
Note that the style Y is not provided by MIT-Adobe-5K but generated using the latest version of Lightroom.
Further, the 3 remaining styles provided by MIT-Adobe-5K, together with the other 5 newly generated auto-retouching styles, are used to test our methods' performance when applied to unseen styles.
All images of these styles are also split into train sets and test sets, but these train sets are not actually involved in training.

\subsection{Single style enhancement}

\begin{figure*}
    \centering
    \!\!\!\!\!
    \subfigure[Input]{
        \includegraphics[width=2.1cm]{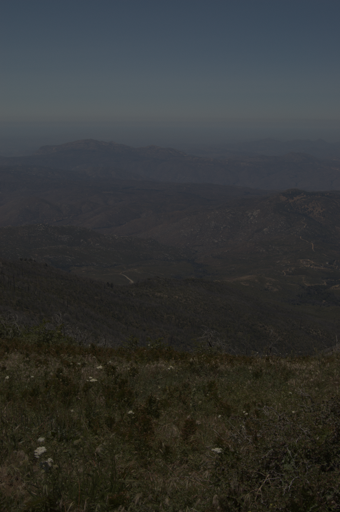}
    }
    \!\!\!\!\!
    \subfigure[DPE~\cite{chen2018deep}]{
        \includegraphics[width=2.1cm]{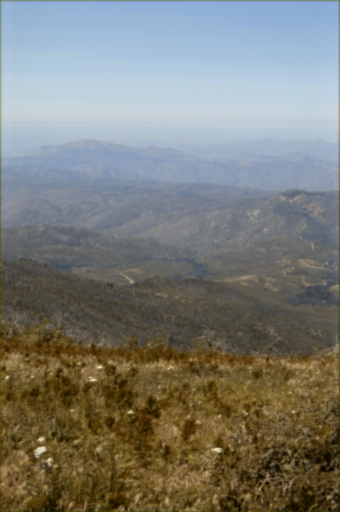}
    }
    \!\!\!\!\!
    \subfigure[HDRNet~\cite{gharbi2017deep}]{
        \includegraphics[width=2.1cm]{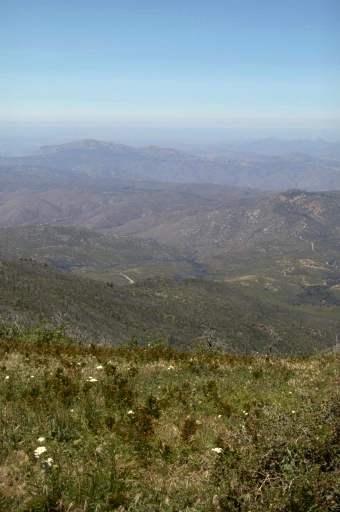}
    }
    \!\!\!\!\!
    \subfigure[DeepUPE~\cite{wang2019underexposed}]{
        \includegraphics[width=2.1cm]{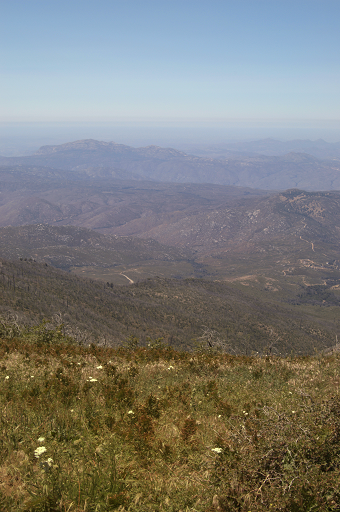}
    }
    \!\!\!\!\!
    \subfigure[DeepLPF~\cite{moran2020deeplpf}]{
        \includegraphics[width=2.1cm]{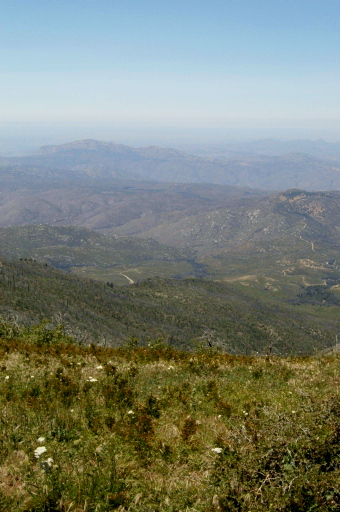}
    }
    \!\!\!\!\!
    \subfigure[A3DLUT~\cite{zeng2020learning}]{
        \includegraphics[width=2.1cm]{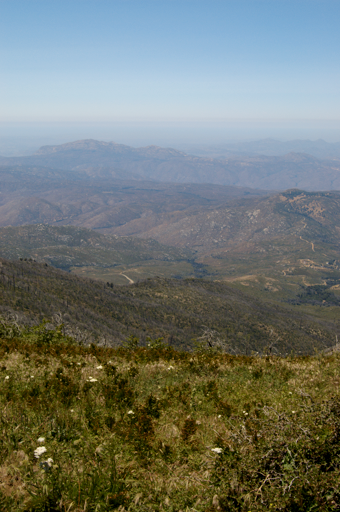}
    }
    \!\!\!\!\!
    \subfigure[StarEnhancer]{
        \includegraphics[width=2.1cm]{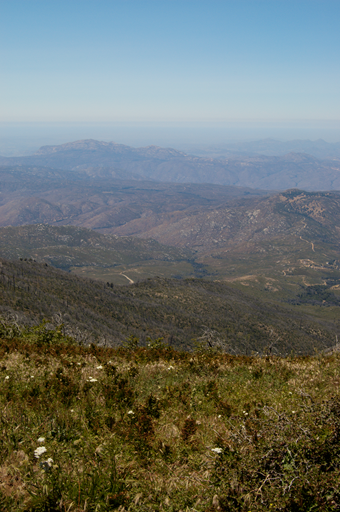}
    }
    \!\!\!\!\!
    \subfigure[Ground Truth]{
        \includegraphics[width=2.1cm]{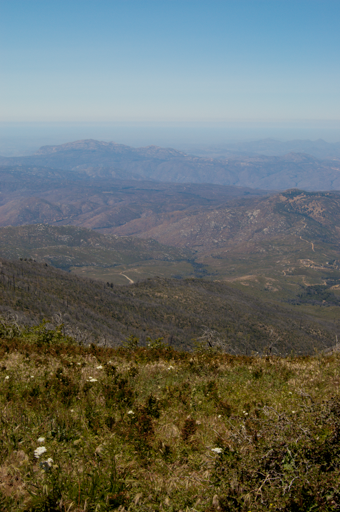}
    }
    \!\!\!\!\!
    \caption{\label{fig:qualitative}
        Qualitative comparison of single style transformation with contemporaneous methods on MIT-Adobe-5K-UPE.
    }
    \!\!\!\!\!
\end{figure*}

\begin{table}[!t]
    \vspace{-0.4 cm}
    \centering
    \caption{
        Quantitative comparison of single style transformation with contemporaneous methods on MIT-Adobe-5K-UPE.
        Running speed (FPS) is measured on 4K-resolution images using a single TITAN RTX.
        Note that some results are replicated from~\cite{moran2020deeplpf,moran2021curl,zeng2020learning}.
    }
    \begin{tabular}{|p{2.1 cm}|p{1.0 cm}<{\centering}|p{1.0 cm}<{\centering}|p{1.0 cm}<{\centering}|p{1.0 cm}<{\centering}|}
        \hline
        \textbf{Method}                     & \textbf{PSNR}     & \textbf{SSIM}     & \textbf{LPIPS}    & \textbf{FPS}                      \\
        \hline\hline
        Exposure~\cite{hu2018exposure}      & 18.57             & 0.701             & --                & 0.11                             \\
        Dis-Rec~\cite{park2018distort}      & 20.97             & 0.841             & --                & 0.009                             \\
        \hline
        U-Net~\cite{ronneberger2015u}       & 22.24             & 0.850             & --                & \multirow{4}{*}{--}               \\
        DPE~\cite{chen2018deep}             & 22.15             & 0.850             & --                &                                   \\
        CURL~\cite{moran2021curl}           & 24.20             & 0.880             & 0.108             &                                   \\
        DeepLPF~\cite{moran2020deeplpf}     & 24.48             & 0.887             & 0.103             &                                   \\
        \hline
        HDRNet~\cite{gharbi2017deep}        & 23.20             & 0.917             & 0.120             & 22                             \\
        DeepUPE~\cite{wang2019underexposed} & 23.24             & 0.893             & 0.158             & 4.7                             \\
        A3DLUT~\cite{zeng2020learning}      & 24.92             & 0.934             & 0.093             & \textbf{602}                     \\
        \hline\hline
        Basic                               & \textbf{25.46}    & \textbf{0.948}    & \textbf{0.083}    & \multirow{2}{*}{\underline{205}} \\
        StarEnhancer                        & \underline{25.29} & \underline{0.943} & \underline{0.086} &                                   \\
        \hline
    \end{tabular}
    \label{tab:quantitative}
\end{table}

We compare our method with contemporaneous methods on MIT-Adobe-5K-UPE as shown in Table~\ref{tab:quantitative}.
StarEnhancer outperforms all the compared methods in terms of PSNR, SSIM, and LPIPS while capable of multi-style enhancement.
Moreover, our proposed enhancer can achieve even better performance if we train the basic enhancer without Dual AdaIN on the unexpanded MIT-Adobe-5K-UPE dataset.
StarEnhancer introduces correlation between channels, which makes it more expressive than another curve-based enhancer named CURL~\cite{moran2021curl}.
Adaptive 3DLUT~\cite{zeng2020learning} achieves excellent performance because of the expressive 3DLUT, but its encoder only predicts fusion weights, which limits its further improvement.
But Adaptive 3DLUT is still the fastest method because it consists of only interpolation and slicing operations with a very lightweight backbone.
Fortunately, StarEnhancer is comparably efficient, and the gap between them is hard to perceive by users.
Note that all U-Net-based methods are unable to enhance 4K-resolution images on a single GPU, which makes them impractical.
Figure~\ref{fig:qualitative} further shows the qualitative comparison results for a single sample.
It can be seen that the image enhanced by StarEnhancer is most similar to the ground truth in both tone and illumination, especially in the sky and grassland.

\subsection{Multi-style enhancement}

\begin{figure}[t]
    \vspace{-0.35 cm}
    \centering
    \includegraphics[width=1.0\columnwidth]{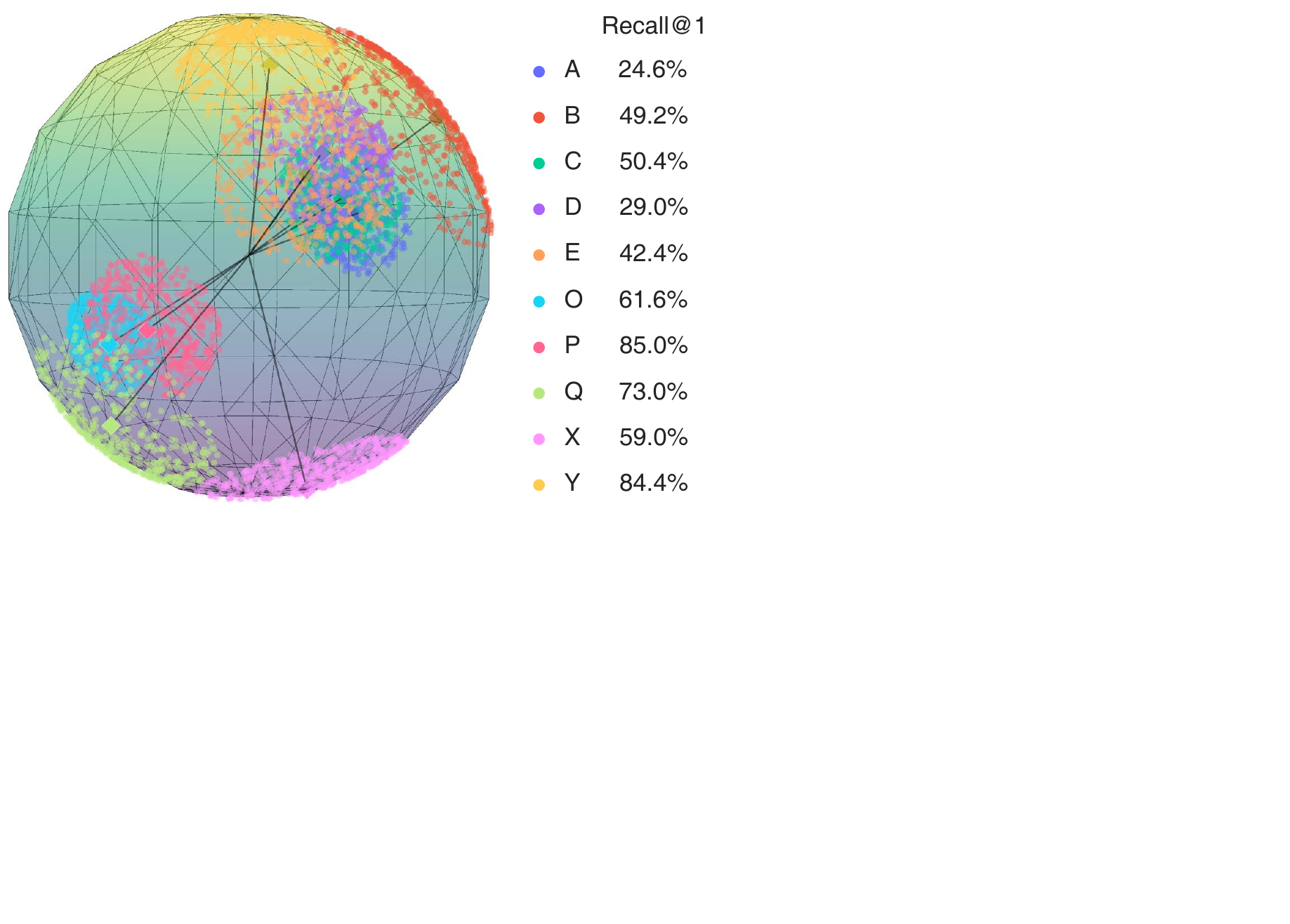}
    \caption{
        Visualization of features learned by style encoder, all features are projected onto the unit sphere using t-SNE~\cite{maaten2008visualizing}.
        Recall@1 for each class on the test set is listed.
    }
    \label{fig:embedding}
\end{figure}

\begin{figure*}[t]
    \centering
    \includegraphics[width=1.0\textwidth]{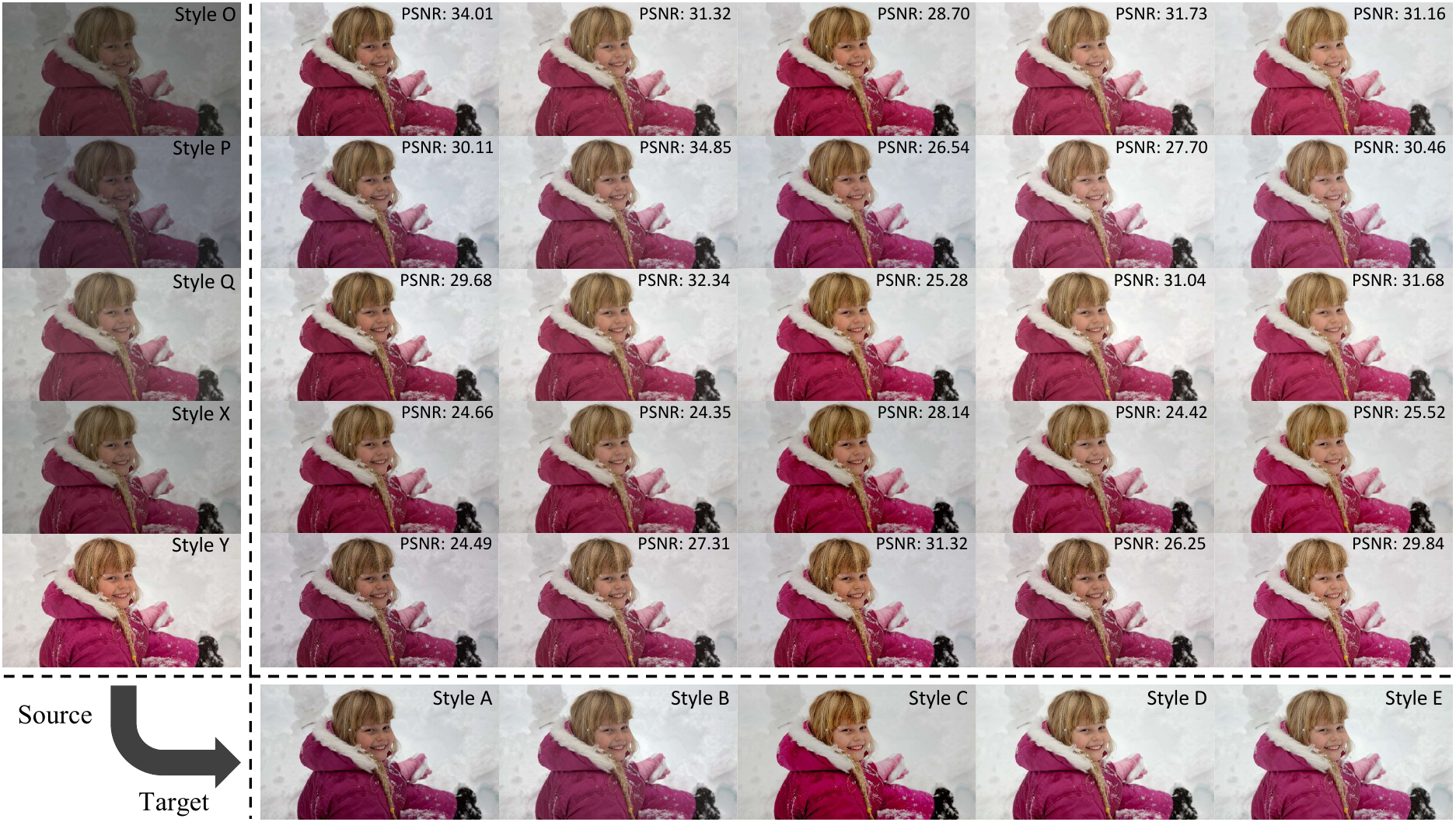}
    \caption{
        An example of mappings between multiple styles.
        The expert retouching styles are used as the target styles, and the other styles are used as the source styles.
    }
    \label{fig:result_4644}
\end{figure*}

\begin{figure}[t]
    \centering
    \includegraphics[width=1.0\columnwidth]{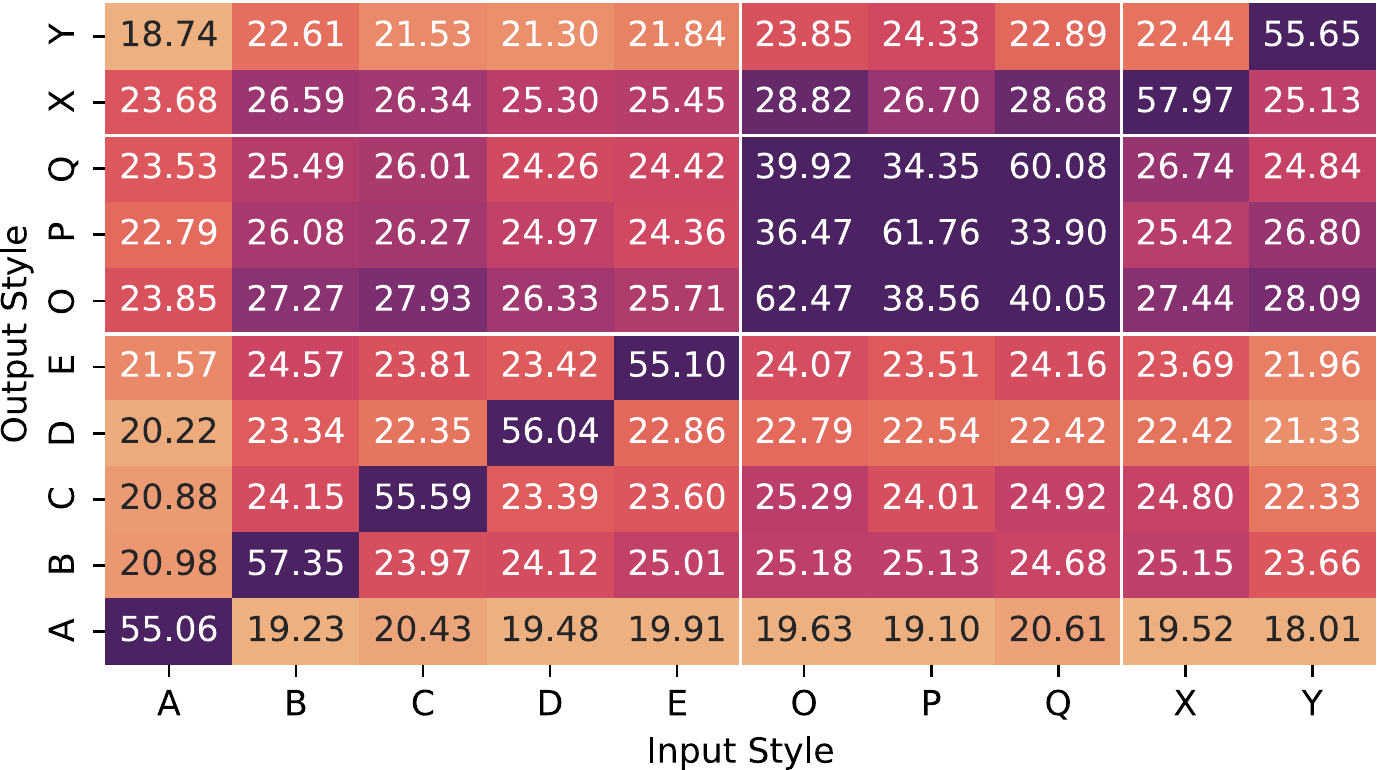}
    \caption{
        Quantitative results of the mappings between multiple styles of the MIT-Adobe-5K dataset~\cite{bychkovsky2011learning} in PSNR.
    }
    \label{fig:all_style}
\end{figure}

We first observe the distribution of features in the embedding space as shown in Figure~\ref{fig:embedding}.
The expert retouching style and the camera input style are distributed on opposite ends of the sphere.
Notably, the auto-retouching style X using the old Lightroom is close to the camera input style, while the auto-retouching style Y using the latest Lightroom is close to the expert retouching style, which can indicate the evolution of auto retouch tools.
Further, Recall@1 illustrates that expert retouching styles are significantly more difficult to distinguish than camera input styles and auto-retouching styles, demonstrating that human aesthetics are subjective and difficult to quantify.

We then focus on evaluating StarEnhancer's multi-style enhancement performance.
Figure~\ref{fig:result_4644} shows some results of the mapping between multiple styles for a single sample.
It is seen that StarEnhancer can adapt to different source styles, even if they vary greatly in brightness and color.
Meanwhile, StarEnhancer can capture the characteristics of the target style to enhance the image.
Specifically, the little girl in learned C's retouched image has more vivid clothes and a more natural-looking face.
Finally, we find that the input style still affects the output image, such as the output images transformed from style P always have a cooler tone.
This may be due to the insufficient variety of scenes in the MIT-Adobe-5K, which prevents the trained StarEnhancer from separating the style information well.

Figure~\ref{fig:all_style} shows the quantitative evaluation results on the multi-style MIT-Adobe-5K benchmark.
We do not add an explicit regular term to the loss function to constrain the same style's transformations, but StarEnhancer still performs impressively.
The transformations between camera input styles are the simplest since only simple global color adjustments (\emph{e.g.} white balance) are applied.
In contrast, transformations between expert retouching styles are much more difficult, and the most difficult style of all is expert retouching style A.
Notably, we believe that the difficulty of learning styles is mainly related to the complexity of their transformations, while the recall of the style encoder indicates mostly the robustness of the transformation.
Specifically, style Y is easier to distinguish but more difficult to learn than style X.
We suppose this is because the new Lightroom's auto retouch tool is more complex and more robust.

\subsection{Functional flexibility}

\begin{figure}[t]
    \centering
    \includegraphics[width=1.0\columnwidth]{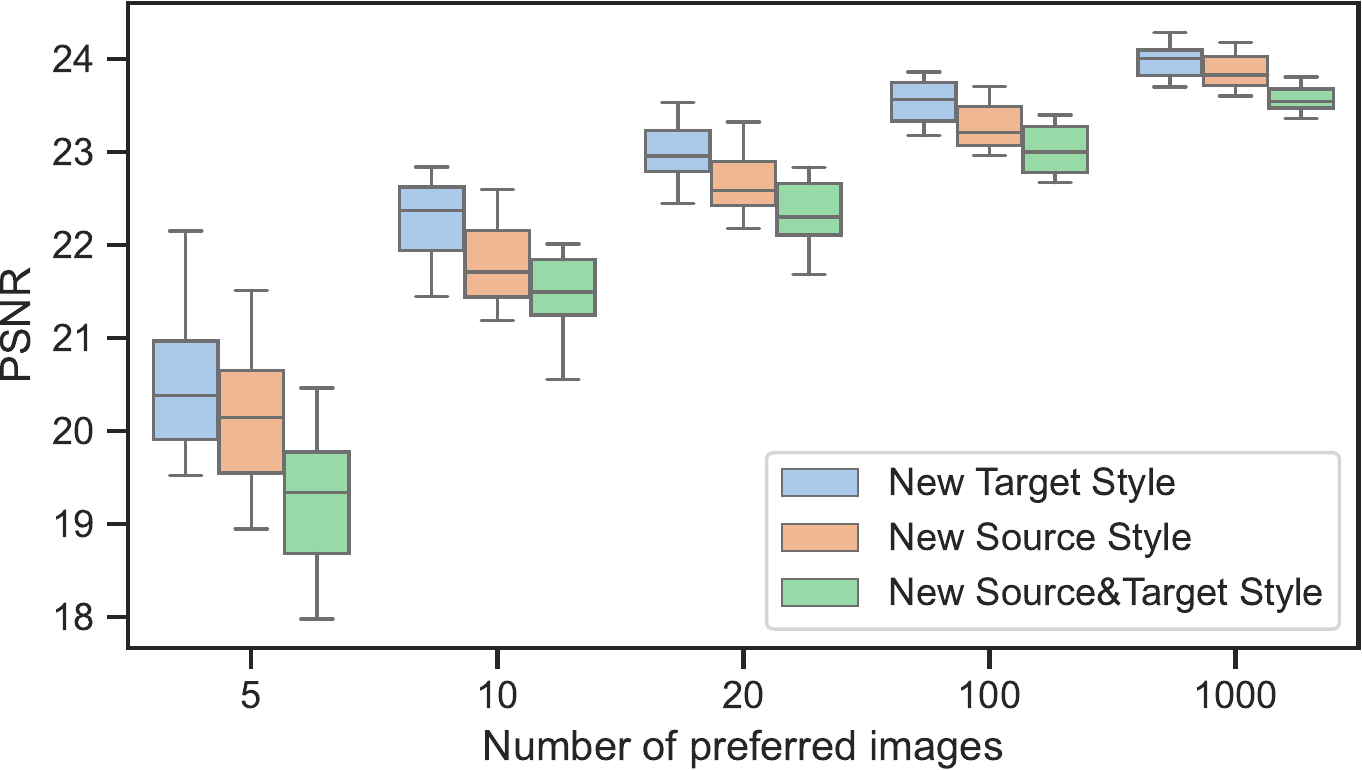}
    \caption{
        Quantitative results when StarEnhancer is applied to unseen styles.
    }
    \label{fig:new_style}
\end{figure}

\begin{figure*}[t]
    \centering
    \includegraphics[width=1.0\textwidth]{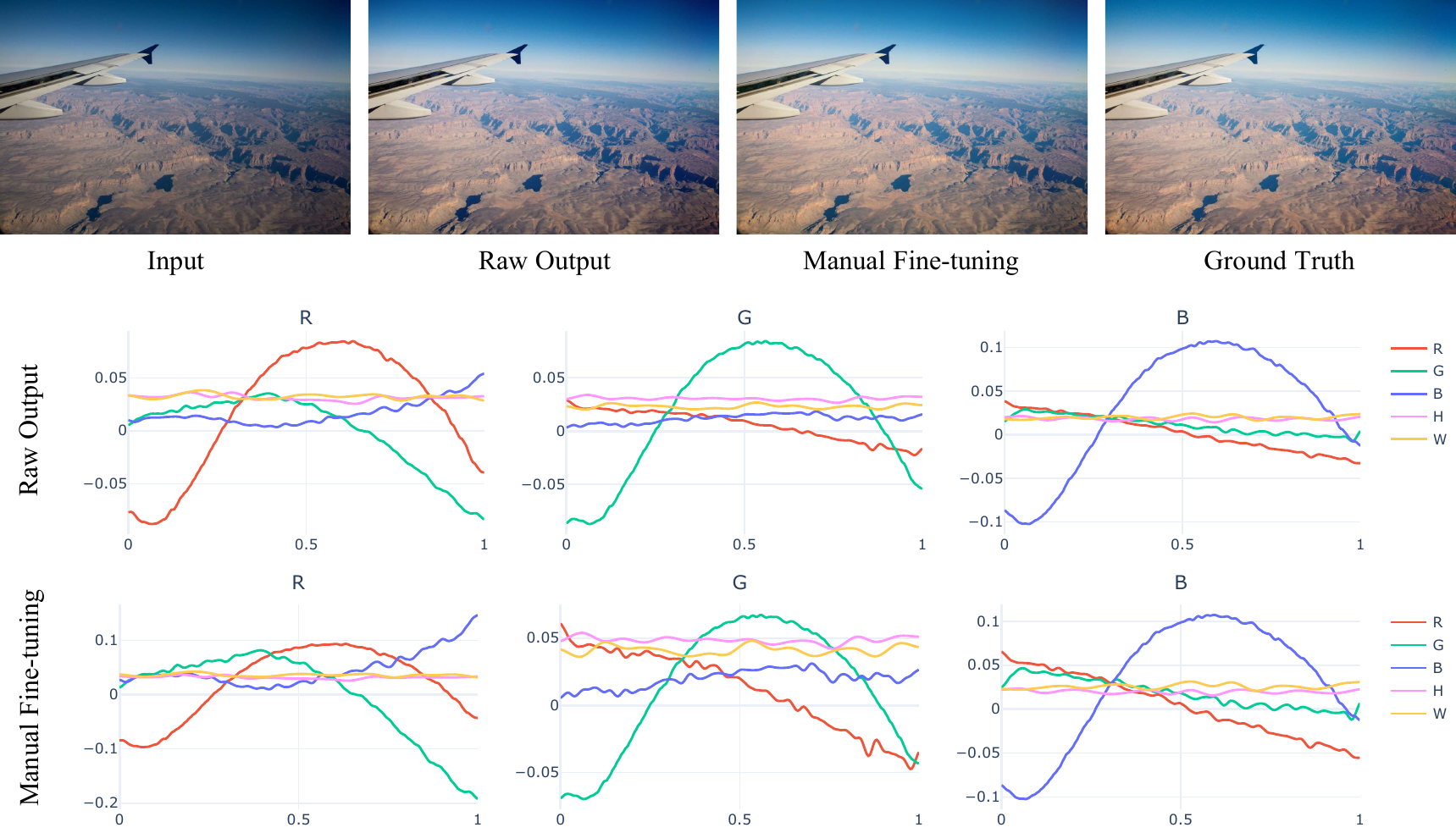}
    \caption{
        An example of manual fine-tuning, where users can obtain a more preferred result with the fine-tuning options.
        The top row lists the input image, the enhanced image output directly from StarEnhancer, the enhanced image after manual fine-tuning, and the ground truth image in the target style.
        The middle row shows the curves predicted by StarEnhancer.
        The bottom row shows the curves adjusted using the slider-based manual fine-tuning tool.
    }
    \label{fig:manual}
\end{figure*}

The style encoder is introduced to enable StarEnhancer to perform enhancements between various unseen styles.
We simulate selecting the users' preferred styles to evaluate starEnhancer's generalization performance using 8 unseen styles.
We randomly choose a source style and a target style and select several image samples to generate new latent codes.
Then we use the mapping network to transform these latent codes into style codes for Dual AdaIN.
Finally, we evaluate the performance of the customized enhancer on the test sets.
Because the number of samples affects the latent codes' reliability, we evaluate the enhancer's generalization performance in different sample size settings and repeat the experiment 10 times at each setting to reduce the error due to the difference in sample selection.
As shown in Figure~\ref{fig:new_style}, the more samples used tend to yield more reliable latent codes, leading to better generalization performance.
And generalizing to an unseen source style is more challenging than generalizing to an unseen target style.
Furthermore, when both the source style and target style are unseen styles, the enhancer's performance decreases further.
But even so, our enhancer can outperform most of the enhancers fine-tuned on the train sets of these styles.

Figure~\ref{fig:manual} shows an example of manual fine-tuning, and each curve's contribution can be observed.
For this sample, the curves that map between the color channels contribute visibly to the residual image.
The curves that map from the pixel's coordinates contribute little to the residual image.
However, we believe that the pixel's coordinates are crucial, especially for some samples that are retouched using the gradient filter or elliptical filter.
Because the enhanced image output directly from StarEnhancer differs significantly from the desired image, we manually adjust each curve's contribution using the proposed fine-tuning tool.
Although the fine-tuning tool only stretches the curves, the fine-tuned image is significantly closer to the desired image.

\section{Conclusion}

In this paper, we propose an expressive curve-based image enhancer that can enhance a 4K-resolution image over 200 FPS.
It surpasses the contemporaneous methods on the MIT-Adobe-5K. 
Based on our proposed style encoder and Dual AdaIN, we extend the enhancer to a multi-style enhancer and name it StarEnhancer, which can perform the mapping between multiple styles using a single model. 
Notably, our proposed approach is flexible enough to be applied to unseen styles. 
Lastly, we introduce a manual fine-tuning tool to meet the user preferences further.

\vspace{0.4 cm}

\noindent
\textbf{Acknowledgement.} This work is supported by National Key Research and Development Program of China under Grant 2020AAA0107400, Zhejiang Provincial Natural Science Foundation of China under Grant No. LZ18F020002, Alibaba-Zhejiang University Joint Research Institute of Frontier Technologies.

\newpage

{\small
\bibliographystyle{ieee_fullname}
\bibliography{egbib}

\begin{thebibliography}{10}\itemsep=-1pt

\bibitem{afifi2020deep}
Mahmoud Afifi and Michael~S Brown.
\newblock Deep white-balance editing.
\newblock In {\em IEEE Conference on Computer Vision and Pattern Recognition
  (CVPR)}, pages 1397--1406, 2020.

\bibitem{afifi2021learning}
Mahmoud Afifi, Konstantinos~G Derpanis, Bjorn Ommer, and Michael~S Brown.
\newblock Learning multi-scale photo exposure correction.
\newblock In {\em IEEE Conference on Computer Vision and Pattern Recognition
  (CVPR)}, pages 9157--9167, 2021.

\bibitem{bickel2020multiple}
Marc Bickel, Samuel Dubuis, and S{\'e}bastien Gachoud.
\newblock Multiple generative adversarial networks analysis for predicting
  photographers' retouching.
\newblock {\em arXiv preprint arXiv:2006.02921}, 2020.

\bibitem{bychkovsky2011learning}
Vladimir Bychkovsky, Sylvain Paris, Eric Chan, and Fr{\'e}do Durand.
\newblock Learning photographic global tonal adjustment with a database of
  input/output image pairs.
\newblock In {\em IEEE Conference on Computer Vision and Pattern Recognition
  (CVPR)}, pages 97--104. IEEE, 2011.

\bibitem{chai2020supervised}
Yoav Chai, Raja Giryes, and Lior Wolf.
\newblock Supervised and unsupervised learning of parameterized color
  enhancement.
\newblock In {\em The IEEE Winter Conference on Applications of Computer Vision
  (WACV)}, pages 992--1000, 2020.

\bibitem{chen2017fast}
Qifeng Chen, Jia Xu, and Vladlen Koltun.
\newblock Fast image processing with fully-convolutional networks.
\newblock In {\em International Conference on Computer Vision (ICCV)}, pages
  2497--2506, 2017.

\bibitem{chen2018deep}
Yu-Sheng Chen, Yu-Ching Wang, Man-Hsin Kao, and Yung-Yu Chuang.
\newblock Deep photo enhancer: Unpaired learning for image enhancement from
  photographs with gans.
\newblock In {\em IEEE Conference on Computer Vision and Pattern Recognition
  (CVPR)}, pages 6306--6314, 2018.

\bibitem{choi2018stargan}
Yunjey Choi, Minje Choi, Munyoung Kim, Jung-Woo Ha, Sunghun Kim, and Jaegul
  Choo.
\newblock Stargan: Unified generative adversarial networks for multi-domain
  image-to-image translation.
\newblock In {\em IEEE Conference on Computer Vision and Pattern Recognition
  (CVPR)}, pages 8789--8797, 2018.

\bibitem{choi2020stargan}
Yunjey Choi, Youngjung Uh, Jaejun Yoo, and Jung-Woo Ha.
\newblock Stargan v2: Diverse image synthesis for multiple domains.
\newblock In {\em IEEE Conference on Computer Vision and Pattern Recognition
  (CVPR)}, pages 8188--8197, 2020.

\bibitem{deng2019arcface}
Jiankang Deng, Jia Guo, Niannan Xue, and Stefanos Zafeiriou.
\newblock Arcface: Additive angular margin loss for deep face recognition.
\newblock In {\em IEEE Conference on Computer Vision and Pattern Recognition
  (CVPR)}, pages 4690--4699, 2019.

\bibitem{fritsch1980monotone}
Frederick~N Fritsch and Ralph~E Carlson.
\newblock Monotone piecewise cubic interpolation.
\newblock {\em SIAM Journal on Numerical Analysis}, 17(2):238--246, 1980.

\bibitem{gharbi2017deep}
Micha{\"e}l Gharbi, Jiawen Chen, Jonathan~T Barron, Samuel~W Hasinoff, and
  Fr{\'e}do Durand.
\newblock Deep bilateral learning for real-time image enhancement.
\newblock {\em ACM Transactions on Graphics (TOG)}, 36(4):1--12, 2017.

\bibitem{goodfellow2014generative}
Ian Goodfellow, Jean Pouget-Abadie, Mehdi Mirza, Bing Xu, David Warde-Farley,
  Sherjil Ozair, Aaron Courville, and Yoshua Bengio.
\newblock Generative adversarial nets.
\newblock {\em Advances in Neural Information Processing Systems (NeurIPS)},
  27:2672--2680, 2014.

\bibitem{grossberg2003determining}
Michael~D Grossberg and Shree~K Nayar.
\newblock Determining the camera response from images: What is knowable?
\newblock {\em IEEE Transactions on Pattern Analysis and Machine Intelligence
  (TPAMI)}, 25(11):1455--1467, 2003.

\bibitem{guo2020zero}
Chunle Guo, Chongyi Li, Jichang Guo, Chen~Change Loy, Junhui Hou, Sam Kwong,
  and Runmin Cong.
\newblock Zero-reference deep curve estimation for low-light image enhancement.
\newblock In {\em IEEE Conference on Computer Vision and Pattern Recognition
  (CVPR)}, pages 1780--1789, 2020.

\bibitem{he2020conditional}
Jingwen He, Yihao Liu, Yu Qiao, and Chao Dong.
\newblock Conditional sequential modulation for efficient global image
  retouching.
\newblock In {\em European Conference on Computer Vision (ECCV)}, pages
  679--695. Springer, 2020.

\bibitem{he2016deep}
Kaiming He, Xiangyu Zhang, Shaoqing Ren, and Jian Sun.
\newblock Deep residual learning for image recognition.
\newblock In {\em IEEE Conference on Computer Vision and Pattern Recognition
  (CVPR)}, pages 770--778, 2016.

\bibitem{howard2017mobilenets}
Andrew~G Howard, Menglong Zhu, Bo Chen, Dmitry Kalenichenko, Weijun Wang,
  Tobias Weyand, Marco Andreetto, and Hartwig Adam.
\newblock Mobilenets: Efficient convolutional neural networks for mobile vision
  applications.
\newblock {\em arXiv preprint arXiv:1704.04861}, 2017.

\bibitem{hu2018exposure}
Yuanming Hu, Hao He, Chenxi Xu, Baoyuan Wang, and Stephen Lin.
\newblock Exposure: A white-box photo post-processing framework.
\newblock {\em ACM Transactions on Graphics (TOG)}, 37(2):1--17, 2018.

\bibitem{huang2017arbitrary}
Xun Huang and Serge Belongie.
\newblock Arbitrary style transfer in real-time with adaptive instance
  normalization.
\newblock In {\em International Conference on Computer Vision (ICCV)}, pages
  1501--1510, 2017.

\bibitem{ioffe2015batch}
Sergey Ioffe and Christian Szegedy.
\newblock Batch normalization: Accelerating deep network training by reducing
  internal covariate shift.
\newblock In {\em International Conference on Machine Learning (ICML)}, pages
  448--456. PMLR, 2015.

\bibitem{jiang2021enlightengan}
Yifan Jiang, Xinyu Gong, Ding Liu, Yu Cheng, Chen Fang, Xiaohui Shen, Jianchao
  Yang, Pan Zhou, and Zhangyang Wang.
\newblock Enlightengan: Deep light enhancement without paired supervision.
\newblock {\em IEEE Transactions on Image Processing (TIP)}, 30:2340--2349,
  2021.

\bibitem{karaimer2016software}
Hakki~Can Karaimer and Michael~S Brown.
\newblock A software platform for manipulating the camera imaging pipeline.
\newblock In {\em European Conference on Computer Vision (ECCV)}, pages
  429--444. Springer, 2016.

\bibitem{kim2020global}
Han-Ul Kim, Young~Jun Koh, and Chang-Su Kim.
\newblock Global and local enhancement networks for paired and unpaired image
  enhancement.
\newblock In {\em European Conference on Computer Vision (ECCV)}, pages
  339--354. Springer, 2020.

\bibitem{kim2020pienet}
Han-Ul Kim, Young~Jun Koh, and Chang-Su Kim.
\newblock Pienet: Personalized image enhancement network.
\newblock In {\em European Conference on Computer Vision (ECCV)}, pages
  374--390. Springer, 2020.

\bibitem{kingma2015adam}
Diederik~P Kingma and Jimmy Ba.
\newblock Adam: A method for stochastic optimization.
\newblock {\em International Conference on Learning Representations (ICLR)},
  2015.

\bibitem{kinoshita2019convolutional}
Yuma Kinoshita and Hitoshi Kiya.
\newblock Convolutional neural networks considering local and global features
  for image enhancement.
\newblock In {\em IEEE International Conference on Image Processing (ICIP)},
  pages 2110--2114. IEEE, 2019.

\bibitem{kneubuehler2020flexible}
Dario Kneubuehler, Shuhang Gu, Luc Van~Gool, and Radu Timofte.
\newblock Flexible example-based image enhancement with task adaptive global
  feature self-guided network.
\newblock In {\em European Conference on Computer Vision (ECCV)}, pages
  343--358. Springer, 2020.

\bibitem{li2020flexible}
Chongyi Li, Chunle Guo, Qiming Ai, Shangchen Zhou, and Chen~Change Loy.
\newblock Flexible piecewise curves estimation for photo enhancement.
\newblock {\em arXiv preprint arXiv:2010.13412}, 2020.

\bibitem{liang2020deep}
Jinxiu Liang, Yong Xu, Yuhui Quan, Jingwen Wang, Haibin Ling, and Hui Ji.
\newblock Deep bilateral retinex for low-light image enhancement.
\newblock {\em arXiv preprint arXiv:2007.02018}, 2020.

\bibitem{liu2020color}
Enyu Liu, Songnan Li, and Shan Liu.
\newblock Color enhancement using global parameters and local features
  learning.
\newblock In {\em Asian Conference on Computer Vision (ACCV)}, 2020.

\bibitem{liu2017sphereface}
Weiyang Liu, Yandong Wen, Zhiding Yu, Ming Li, Bhiksha Raj, and Le Song.
\newblock Sphereface: Deep hypersphere embedding for face recognition.
\newblock In {\em IEEE Conference on Computer Vision and Pattern Recognition
  (CVPR)}, pages 212--220, 2017.

\bibitem{long2015fully}
Jonathan Long, Evan Shelhamer, and Trevor Darrell.
\newblock Fully convolutional networks for semantic segmentation.
\newblock In {\em IEEE Conference on Computer Vision and Pattern Recognition
  (CVPR)}, pages 3431--3440, 2015.

\bibitem{lore2017llnet}
Kin~Gwn Lore, Adedotun Akintayo, and Soumik Sarkar.
\newblock Llnet: A deep autoencoder approach to natural low-light image
  enhancement.
\newblock {\em Pattern Recognition (PR)}, 61:650--662, 2017.

\bibitem{loshchilov2016sgdr}
Ilya Loshchilov and Frank Hutter.
\newblock Sgdr: Stochastic gradient descent with warm restarts.
\newblock In {\em International Conference on Learning Representations (ICLR)},
  2017.

\bibitem{maaten2008visualizing}
Laurens van~der Maaten and Geoffrey Hinton.
\newblock Visualizing data using t-sne.
\newblock {\em Journal of Machine Learning Research (JMLR)}, 9(Nov):2579--2605,
  2008.

\bibitem{moran2020deeplpf}
Sean Moran, Pierre Marza, Steven McDonagh, Sarah Parisot, and Gregory Slabaugh.
\newblock Deeplpf: Deep local parametric filters for image enhancement.
\newblock In {\em IEEE Conference on Computer Vision and Pattern Recognition
  (CVPR)}, pages 12826--12835, 2020.

\bibitem{moran2021curl}
Sean Moran, Steven McDonagh, and Gregory Slabaugh.
\newblock Curl: Neural curve layers for global image enhancement.
\newblock In {\em International Conference on Pattern Recognition (ICPR)},
  pages 9796--9803. IEEE, 2021.

\bibitem{park2018distort}
Jongchan Park, Joon-Young Lee, Donggeun Yoo, and In So~Kweon.
\newblock Distort-and-recover: Color enhancement using deep reinforcement
  learning.
\newblock In {\em IEEE Conference on Computer Vision and Pattern Recognition
  (CVPR)}, pages 5928--5936, 2018.

\bibitem{paszke2019pytorch}
Adam Paszke, Sam Gross, Francisco Massa, Adam Lerer, James Bradbury, Gregory
  Chanan, Trevor Killeen, Zeming Lin, Natalia Gimelshein, Luca Antiga, et~al.
\newblock Pytorch: An imperative style, high-performance deep learning library.
\newblock {\em Advances in Neural Information Processing Systems (NeurIPS)},
  32:8026--8037, 2019.

\bibitem{ronneberger2015u}
Olaf Ronneberger, Philipp Fischer, and Thomas Brox.
\newblock U-net: Convolutional networks for biomedical image segmentation.
\newblock In {\em International Conference on Medical Image Computing and
  Computer-Assisted Intervention (MICCAI)}, pages 234--241. Springer, 2015.

\bibitem{tan2019efficientnet}
Mingxing Tan and Quoc Le.
\newblock Efficientnet: Rethinking model scaling for convolutional neural
  networks.
\newblock In {\em International Conference on Machine Learning (ICML)}, pages
  6105--6114. PMLR, 2019.

\bibitem{tao2017llcnn}
Li Tao, Chuang Zhu, Guoqing Xiang, Yuan Li, Huizhu Jia, and Xiaodong Xie.
\newblock Llcnn: A convolutional neural network for low-light image
  enhancement.
\newblock In {\em IEEE Visual Communications and Image Processing (VCIP)},
  pages 1--4. IEEE, 2017.

\bibitem{wang2017normface}
Feng Wang, Xiang Xiang, Jian Cheng, and Alan~Loddon Yuille.
\newblock Normface: L2 hypersphere embedding for face verification.
\newblock In {\em ACM International Conference on Multimedia (ACM MM)}, pages
  1041--1049, 2017.

\bibitem{wang2018cosface}
Hao Wang, Yitong Wang, Zheng Zhou, Xing Ji, Dihong Gong, Jingchao Zhou, Zhifeng
  Li, and Wei Liu.
\newblock Cosface: Large margin cosine loss for deep face recognition.
\newblock In {\em IEEE Conference on Computer Vision and Pattern Recognition
  (CVPR)}, pages 5265--5274, 2018.

\bibitem{wang2019underexposed}
Ruixing Wang, Qing Zhang, Chi-Wing Fu, Xiaoyong Shen, Wei-Shi Zheng, and Jiaya
  Jia.
\newblock Underexposed photo enhancement using deep illumination estimation.
\newblock In {\em IEEE Conference on Computer Vision and Pattern Recognition
  (CVPR)}, pages 6849--6857, 2019.

\bibitem{wei2018deep}
Chen Wei, Wenjing Wang, Wenhan Yang, and Jiaying Liu.
\newblock Deep retinex decomposition for low-light enhancement.
\newblock {\em British Machine Vision Conference (BMVC)}, 2018.

\bibitem{zamir2020learning}
Syed~Waqas Zamir, Aditya Arora, Salman Khan, Munawar Hayat, Fahad~Shahbaz Khan,
  Ming-Hsuan Yang, and Ling Shao.
\newblock Learning enriched features for real image restoration and
  enhancement.
\newblock In {\em European Conference on Computer Vision (ECCV)}, pages
  492--511. Springer, 2020.

\bibitem{zeng2020learning}
Hui Zeng, Jianrui Cai, Lida Li, Zisheng Cao, and Lei Zhang.
\newblock Learning image-adaptive 3d lookup tables for high performance photo
  enhancement in real-time.
\newblock {\em IEEE Transactions on Pattern Analysis and Machine Intelligence
  (TPAMI)}, 2020.

\bibitem{zhang2018fixup}
Hongyi Zhang, Yann~N Dauphin, and Tengyu Ma.
\newblock Fixup initialization: Residual learning without normalization.
\newblock In {\em International Conference on Learning Representations (ICLR)},
  2018.

\bibitem{zhang2018unreasonable}
Richard Zhang, Phillip Isola, Alexei~A Efros, Eli Shechtman, and Oliver Wang.
\newblock The unreasonable effectiveness of deep features as a perceptual
  metric.
\newblock In {\em IEEE Conference on Computer Vision and Pattern Recognition
  (CVPR)}, pages 586--595, 2018.

\bibitem{zhang2018shufflenet}
Xiangyu Zhang, Xinyu Zhou, Mengxiao Lin, and Jian Sun.
\newblock Shufflenet: An extremely efficient convolutional neural network for
  mobile devices.
\newblock In {\em IEEE Conference on Computer Vision and Pattern Recognition
  (CVPR)}, pages 6848--6856, 2018.

\bibitem{zhang2019kindling}
Yonghua Zhang, Jiawan Zhang, and Xiaojie Guo.
\newblock Kindling the darkness: A practical low-light image enhancer.
\newblock In {\em ACM International Conference on Multimedia (ACM MM)}, pages
  1632--1640, 2019.

\end{thebibliography}
}

\end{document}